\documentclass[lettersize,journal]{IEEEtran}

\usepackage[backend=bibtex,style=ieee,natbib=true]{biblatex}
\usepackage{hyperref}
\usepackage{todonotes}

\begin{document}

\title{Intelligent Road Condition Monitoring using 3D In-Air SONAR Sensing}
\author{Amber Cassimon, Robin Kerstens, Walter Daems, Jan Steckel*\thanks{All authors are with the Department of Electronics and ICT Engineering Technology, Cosys-Lab Research Group, University of Antwerp, Antwerp, Belgium\; Flanders Make Strategic Research Centre, Lommel, Belgium}\thanks{*Corresponding Author: \href{mailto:jan.steckel@uantwerpen.be}{jan.steckel@uantwerpen.be}}}


\markboth{Intelligent Road Condition Monitoring using 3D In-Air SONAR Sensing}{Cassimon \textsc{et al.}}



\maketitle

\begin{abstract}
In this paper, we investigate the capabilities of in-air 3D SONAR sensors for the monitoring of road surface conditions.
Concretely, we consider two applications: Road material classification and Road damage detection and classification.
While such tasks can be performed with other sensor modalities, such as camera sensors and LiDAR sensors, these sensor modalities tend to fail in harsh sensing conditions, such as heavy rain, smoke or fog.
By using a sensing modality that is robust to such interference, we enable the creation of opportunistic sensing applications, where vehicles performing other tasks (garbage collection, mail delivery, etc.) can also be used to monitor the condition of the road.
For these tasks, we use a single dataset, in which different types of damages are annotated, with labels including the material of the road surface.
In the material classification task, we differentiate between three different road materials: Asphalt, Concrete and Element roads.
In the damage detection and classification task, we determine if there is damage, and what type of damage (independent of material type), without localizing the damage.
We are succesful in determining the road surface type from SONAR sensor data, with F1 scores approaching 90\% on the test set, but find that for the detection of damages performace lags, with F1 score around 75\%.
From this, we conclude that SONAR sensing is a promising modality to include in opportunistic sensing-based pavement management systems, but that further research is needed to reach the desired accuracy.
\end{abstract}




\section{Introduction}
\label{sec:intro}
    \IEEEPARstart{P}{avement} management systems (PMS) allow municipalities to plan and execute road maintenance in an intelligent fashion~\cite{Shon_2022_Autonomous}.
    However, such PMS can only fulfill their task if sufficient high-quality data is available to accurate assess the state of the road surface.
    Traditionally, data on road surface conditions are gathered manually, i.e., the municipality sends out a worker to a specific location to assess the state of a road.
    This process is time-consuming for the worker, and expensive for the municipality, while yielding only very limited coverage of a municipality's road network.
    This process has been partially automated through the use of dedicated mapping vehicles.
	These mapping vehicles are equipped with a state-of-the-art sensor suite, and continuously drive around, collecting information while driving.
    This improves the coverage of the data being sent to the PMS over the manual approach, but still has some drawbacks.
	In municipalities with large road networks, individual roads are often still revisited at a relatively low frequency, since such mapping vehicles need to cover the entire municipalities road network.
	Additionally, for smaller municipalities with smaller income streams, the acquisition, operation and maintenance of dedicated mapping vehicles may be prohibitively expensive~\cite{Shon_2022_Autonomous}.
    An alternative to dedicated mapping vehicles that resolves these issues is the use of opportunistic sensing~\cite{Hu_2023_Forecasting}.
	In opportunistic sensing, a set of sensors is mounted on vehicles which drive around for other purposes, such as mail delivery, garbage collection, etc.
	These vehicles, by virtue of their purpose, already cover most of a municipality's road network at relatively frequent intervals.
	This makes them excellent vehicles platforms for the collection of road damage data to feed into a PMS.
    Given the size of fleets of vehicles often used for these tasks, it is imperative that the sensors mounted on such vehicles have a low cost, so they can be installed on as many vehicles as possible, to ensure sufficient coverage.
    It is this setting that forms the context for this paper.
	We propose a road surface condition monitoring system based around the use of 3D in-air SONAR (Sound Navigation and Ranging) sensors.
	We consider the task of identifying the material of the road surface, as well as the task of detecting the presence and type of road surface defects.
	Our results show that we can succesfully detect road surface type with 3D in-air SONAR sensors, while we show that damage detection is viable, though its performance lags that of road surface type detection. Finally, we propose several possible avenues for future improvements.

    \noindent The remainder of this paper is structured as follows: Section~\ref{sec:related_work} will look at related work for both road material identification and road damage detection.
	Next, in section~\ref{sec:methodology} we explain our methodology for both tasks.
	Section~\ref{sec:experiments} explains the exact experimental set-up we use for our experiments, and section~\ref{sec:discussion} looks at the results of these experiments.
	Finally, we provide some closing thoughts on the work done in section~\ref{sec:conclusion}, along with some possible future avenues for improvements.

\section{Related Work}
\label{sec:related_work}
	\noindent In this section, we give a brief overview of some relevant works, trying to showcase the breadth of the existing body of literature, without trying to exhaustively enumerate all the work in this area.
	We consider both the problem of detecting and identifying damages in road surfaces, as well as identifying the material of the road surface.

    \subsection{Road Damage Detection}
		\noindent In their work,~\cite{Trinh2024Improving} make use of an RGB (Red, Green, Blue) camera to assess the quality of a road surface into one of three quality categories: bad, regular and good.
		Specifically, their research focuses on the use of a segmentation model to isolate the road from other parts of the camera image.
		They show that the addition of a masking operation with a mask generated by a segmentation model improves classification performance in all classes.
        
        Shim et al.~\cite{Shim_2021_Road} make use of a lightweight auto-encoder network to identify road surface damage based on camera images.
		While they are capable of detecting the presence of road surface damages, they do not consider the identification of individual damages, nor do they try to make an assessment of the overall quality of the road.
        While both Shim et al., and Trinh et al. made use of camera images for their assessment of road damages, other sensing modalities have also been used.
        
        Li et al.~\cite{Li_2025_Research} make use of tire noise to identify pavement damages, they consider pavement to be in normal condition, to exhibit crack damages or to exhibit pothole damages.
		They achieve an overall accuracy of 88.4\% using a random forest combined with gradient boosting.
        
        The accelerometer sensors included in smartphones, combined with GPS (Global Positioning System) data were used by Dong et al.~\cite{Dong_2021_Smartphone} to identify road surface distortions, patching, potholes and rutting.
		Data collection happened on a moving vehicle. Using k-means clustering, they were able to achieve an average accuracy of 84\%.
        
        Ultrasound sensors have also been used to identify damages in concrete structures, such as in the work of Kim et al~\cite{Kim_2023_Initial}.
		They studied concrete specimens in a laboratory environment to identify various characteristics caused by freeze-thaw cycles experienced by the concrete specimens.
		While this work focused on damage detection in concrete, unlike the others, it does not focus on a vehicle-borne sensing case.

    \subsection{Road Material Identification}
    	\noindent Previous work has been done on the use of visual-tactile sensor fusion to identify different road materials and surfaces~\cite{Shi_2023_CNN}.
		In their work, Shi et al. perform sensor fusion between a camera and tactile information derived from a PVDF (Polyvinylidene fluoride) sensor.
		They fuse these sensors using transformer neural networks, and obtain strong predictive performance across all four considered classes.
    	
    	Ultrasound sensing has also been used to identify road surface materials before.
		Kim et al.~\cite{Kim_2021_Road} mount ultrasonic sensors in front of the front wheels of a vehicle, and use a short-time fourier transform coupled with a deep neural network to determine the type of road surface currently being driven on.
		Across 8 road surface types, they achieve accuracies upwards of 95\%.
    	
    	Sattar et al.~\cite{Sattar_2021_Developing} test the use of both RADAR (Radio Detection and Ranging) and SONAR to classify five different road surface types.
		They test both 24GHz RADAR and 40kHz SONAR separately, as well as combined.
		They also consider the use of 150GHz RADAR. They show accuracies of upward of 80\% using both RADAR and SONAR, with the combined results going up to 92\% accuracy.
		While our work only considers on-road surfaces, this work also included ``grass'' as one of the road surface types.


\section{Methodology}
\label{sec:methodology}
    \noindent We consider both the surface material classification, as classification problems, omitting the added complexities of locating the damages, or assessing their severity.
    We always view this through the lens of a multilabel classification problem, since it is possible that multiple damages are captured in one sample, as well as multiple materials, when a road changes from one material type to another.
    We consider a variety of models, including a linear model (logistic regression), a support vector machine, a random forest, a gradient-boosted tree, a decision tree,  and a multi-layer perceptron.
	Additionally, given its additional complexity, we also consider a CNN (Convolutional Neural Network)-based classifier for the damage detection problem, but not for the material identification problem.
    This gives us a set of models covering a wide range of complexities, allowing us to select the simplest model that fits the data well.
    The SONAR data collected for this paper was collected using an eRTIS (Embedded Real-Time Imaging) SONAR array sensor~\cite{Laurijssen2025Ruggedized,Kerstens2019eRTIS,Steckel2012ASonar}.
    The eRTIS is a fully embedded 32-channel 3D in-air SONAR sensor, capable of gathering data under harsh sensing conditions, such as in the presence of fog or dust, where visual sensing methods may struggle.

	\begin{figure}
		\centering
		\includegraphics[width=0.8\columnwidth]{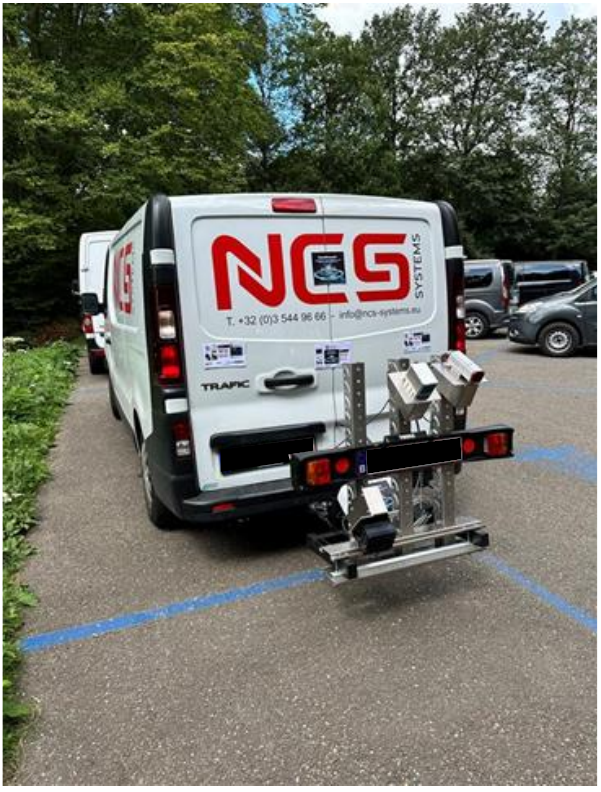}
		\caption{An image showing the mounting of the eRTIS SONAR sensor (bottom left of the sensorbox) on a vehicle. The sensorbox is mounted on the back of the vehicle, looking behind.}
		\label{fig:sensorbox-mounted-on-car}
	\end{figure}

	\begin{figure}
		\centering
		\includegraphics[width=\columnwidth]{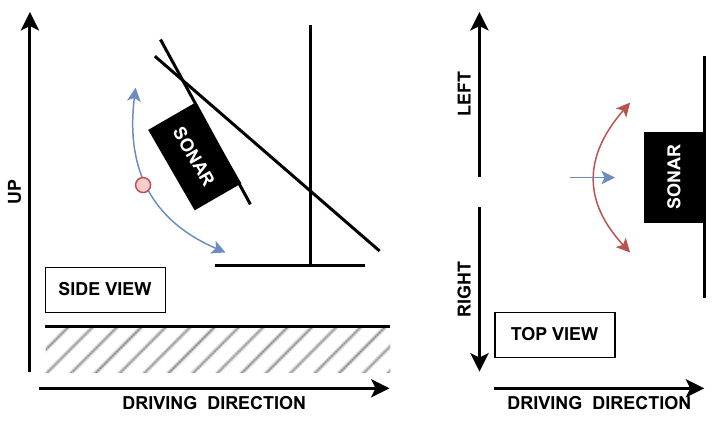}
		\caption{A diagram showing how the SONAR sensor was mounted to the vehicle. The range of elevation angles is indicated in red, and the range of azimuth angles in blue.}
		\label{fig:sonar-mounting-diagram}
	\end{figure}

	Figure~\ref{fig:sensorbox-mounted-on-car} shows the mounting of the sensor on the car. Note that the sensor is mounted on its side. This is further clarified in Figure~\ref{fig:sonar-mounting-diagram}, showing a side view and a top view of the mounting of the  SONAR sensor in a diagram. The range of elevation angles is indicated in red, while the range of azimuth angles is indicated in blue. Due to the way the sensor is mounted, these angles actually differ from what would intuitively be expected, with the elevation angle sweeping from left to right from the driver's point of view, and the azimuth angle sweeping up and down.

    \subsection{Dataset}
    \label{subsec:methodology:dataset}
        \noindent We used a newly collected dataset.
		This dataset contains camera images, raw PDM (Pulse Density Modulation) datastreams recorded by the SONAR sensor, and labels for each camera image.
		Other datapoints were also collected, but excluded from this work. Camera images were only used for labeling, and were not used in the actual detection or classification of damages.
        First, both data types (SONAR recording and labels) are time-synchronized.
		Since these are sampled at different sample rates, we pick one data type (labels) as our reference point, and then select the temporally closest samples from the other sensors.
		Note that this may include a sensor sample that was recorded before the camera image used to label the damage.
        Next, we asses the synchronization of each of the samples.
		If the time-difference between the labels and the other modalities is too high (Above 150ms in our case), the sample is discarded.
		Given that our SONAR sensor operates at a sampling frequency of around 10Hz, this mainly occurs in cases where one sensor was disabled or failed.
        Following synchronization, the material types present in a sample are determined, and this annotation is added to the dataset.
		Note that, during the annotation process, labels were assigned which include both the material, and the damage type. For instance: ``Asphalt - Alligator Crack'' indicates that an alligator crack was found in an asphalt surface.
		The different material types considered are shown in Figure~\ref{fig:material-distribution}, while the different types of damages are shown in Figure~\ref{fig:damage-distribution}.
        Following this, similar damages in different material types are combined into one damage type, shared among the different materials.
		For instance, the labels ``Asphalt - Alligator Crack'' and ``Concrete - Alligator Crack'' are combined into one label ``Alligator Crack'', discarding the material type, which was determined and separately annotated earlier.
        After this, we perform a minimal filtering of the classes considered, eliminating classes that we don't have enough data for.
		Any classes that we have less than 100 samples for are eliminated, to reduce dataset imbalanbce, and prevent classes with very few samples from affecting the overall accuracy.
		We keep this threshold as low as possible, since the utility of a damage classification systems is reduced significantly as the number of detectable damages is reduced.
		Of course, this needs to be weighed against overall classifier performance, given that the collection and labelling of additional samples is costly.
        Additionally, it must be noted that the data was annotated based on camera images.
		This means that the distribution of labels will likely be skewed towards damages that are easily visually detectable, while classes which are harder to detect visually may be underrepresented due to the labelling process.
        Finally, we split the dataset. We first isolate 10\% of the dataset as a held out test set.
		The remaining 90\% of the dataset is split into 10 folds, each with a 90\%-10\% split between training and validation data, respectively.
		Splitting the data into separate folds is done in a stratified fashion, to ensure each fold maximally represents the dataset as a whole.
		Stratification is done based on a ``fake'' feature, that represents a sample's material or damage type, depending on the problem at hand.
		Since we treat the classification as a multilabel classification problem, each combination of damages or materials is treated as a separate value for this feature.
		The feature is built by taking the one-hot encoded vector of the present labels, and multiplying each bit with increasing powers of two, essentially interpreting the one-hot encoded vector as a binary-encoded integer.
		This integer value is used for stratification.

        \begin{figure}
        	\centering
        	\includegraphics[width=.8\columnwidth]{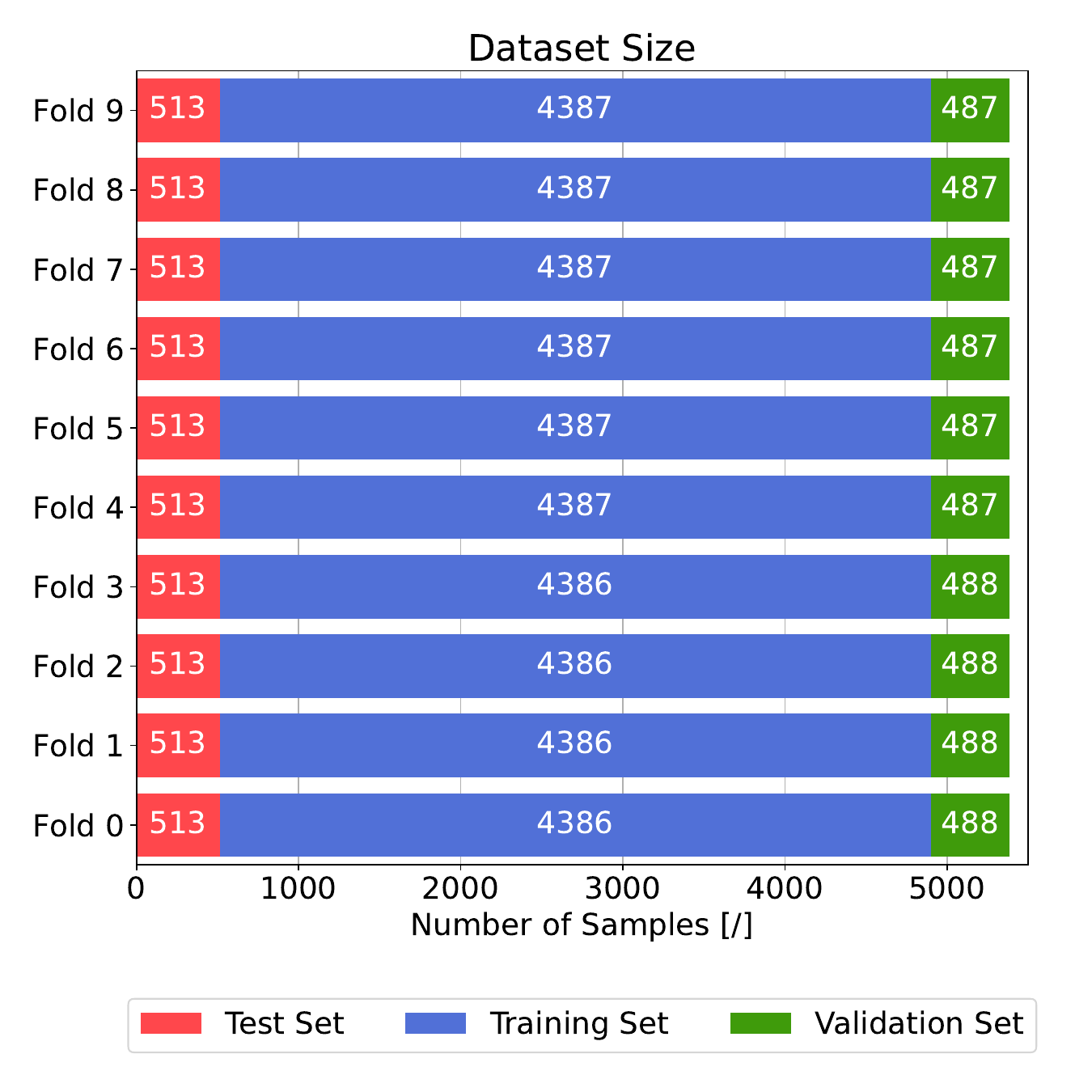}
        	\caption{The size of each dataset, for each fold in the dataset.}
        	\label{fig:fold_sizes}
        \end{figure}

        The size of the different datasets (train, test and validation) across the cross-validation folds is shown in Figure~\ref{fig:fold_sizes}.
        The distribution of labels in the dataset is shown in Figure~\ref{fig:damage-distribution} for the damage types, and Figure~\ref{fig:material-distribution} for the materials.
        Note that labels aren't mutually exclusive in both figures, i.e., one sample can be given more than one label, so the number of samples per class need not sum to the total number of samples.

        \begin{figure}
        	\centering
        	\includegraphics[width=\columnwidth]{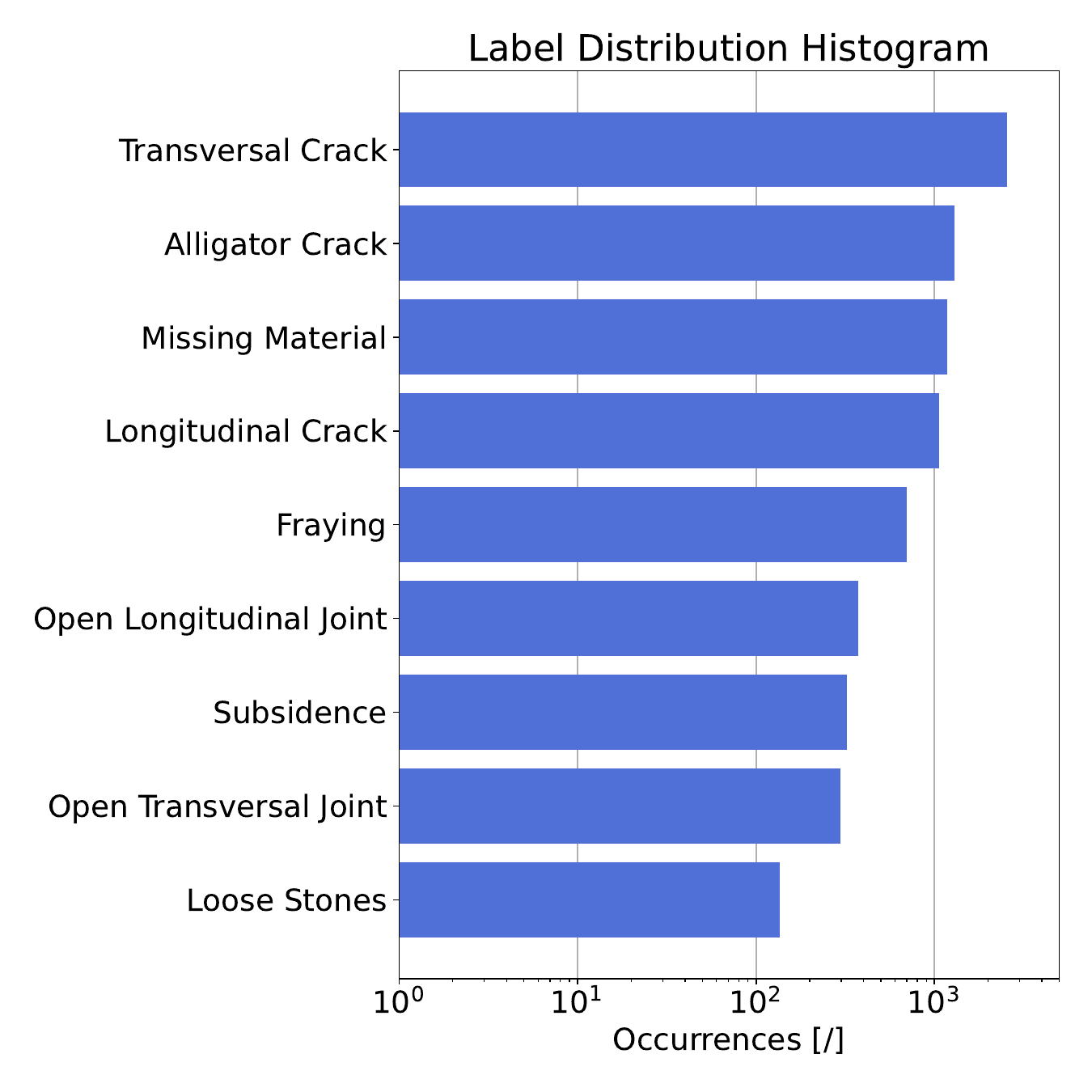}
        	\caption{The distribution among different damages in the dataset. Note the logarithmic X-axis.}
        	\label{fig:damage-distribution}
        \end{figure}

    	\begin{figure}
    		\centering
    		\includegraphics[width=\columnwidth]{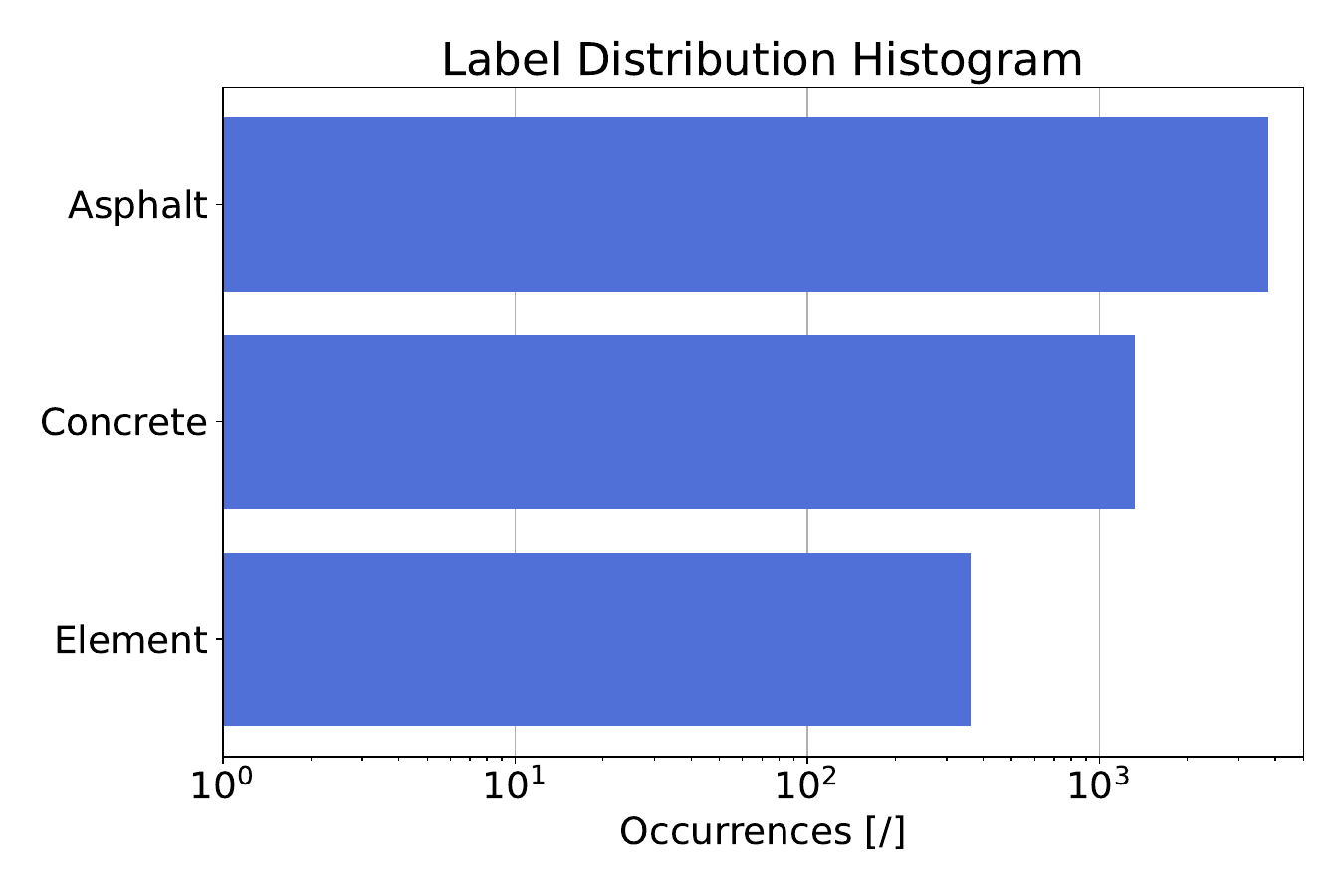}
    		\caption{The distribution among different materials in the dataset. Note the logarithmic X-axis.}
    		\label{fig:material-distribution}
    	\end{figure}
        
    \subsection{Preprocessing}
    \label{subsec:methodology:preprocessing}

    	\begin{figure*}
    		\centering
    		\includegraphics[width=\textwidth]{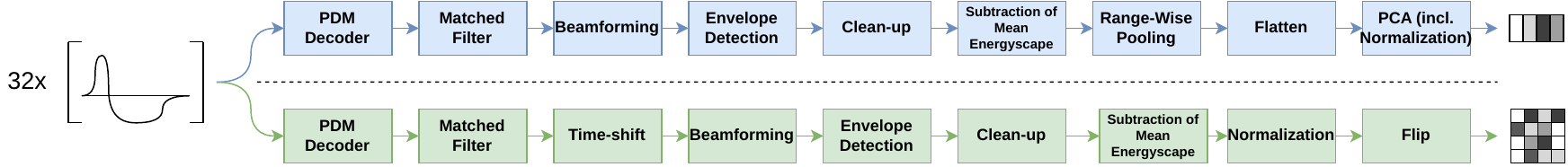}
    		\caption{The vector preprocessing pipeline (top, blue) and the image preprocessing pipeline (below, green). Both pipelines are fed using time-domain, PDM-encoded SONAR recordings for all channels, and output either a feature vector or energyscape, depending on the pipeline.}
    		\label{fig:preprocessing-pipeline}
    	\end{figure*}

        \noindent To prepare the data for consumption by the machine learning models, we employ two different preprocessing pipelines.
		One is used for the CNN model, while the other is used for all other models.
		We make use of a separate pipeline for the CNN model, since it is capable of making use of the spatial relation between closely grouped data points, while remaining independent of the overall position of each data point in the energyscape.
		The pipeline used for the CNN model produces an image and we will thus refer to it as the image preprocessing pipeline, while the other preprocessing pipeline produces a vector, and we will refer to it as the vector preprocessing pipeline.
		We graphically show the different steps in both pipelines in Figure~\ref{fig:preprocessing-pipeline}.

        \subsubsection{Vector preprocessing pipeline}
        \label{subsubsec:methodology:preprocessing:baseline}
            \noindent When preprocessing the data, we make use of the usual signal processing pipeline that is used in conjunction with the eRTIS SONAR sensor~\cite{Steckel2012ASonar}.
			This pipeline consists of a PDM decoder, a matched filter, a delay-and-sum beamformer, an envelope detector and a clean-up stage.
			The clean-up stage works similar to the constant false-alarm rate detection algorithm.
			This signal processing pipeline generates a matrix representing the intensity of the return received from a particular angle (azimuth and elevation) and range.
			This matrix is referred to as an energyscape. Next, we subtract the mean energyscape across all training samples from each of the energyscapes, to emphasize the difference between individual energyscapes.
			We apply a max-pooling filter to the energyscape along the range direction to reduce the dimensionality of the samples.
			Finally, we flatten the energyscape into a vector by concatenating each of the rows, and use Principal Component Analysis (PCA) to reduce its dimensionality.
			It is this set of principal components of the flattened energyscape that is presented as input data to the models.

        \subsubsection{Image preprocessing pipeline}
        \label{subsubsec:methodology:preprocessing:deep_learning}
        	\noindent For the CNN model, we employ a different preprocessing pipeline, to allow the CNN to exploit the data locality inherent in the image-based representation of an energyscape.
            First, we split the signal processing pipeline for the raw SONAR data into two separate pipelines.
			The first includes PDM decoding and matched filtering, while the second includes beamforming, envelope detection and clean-up.
            We start, as we did for the other models, by performing PDM decoding and matched filtering.
			However, before we pass the filtered signals to the beamformer, we include a data augmentation to improve the range-invariance of the model.
			We achieve this by shifting the filtered signals forward or backward by a random amount of samples. Samples that are shifted out are replaced by zeros.
			To ensure that we don't lose any important information, this time-shift is applied to all 32 microphone signals equally.
			After this augmentation, we perform beamforming, envelope detection and energyscape clean-up.
            After signal processing, similar to the other models, we subtract the mean energyscape, computed across all energyscapes.
			We also compute the mean and standard deviation of an individual energyscape, and use this to normalize the individual energyscape by subtracting the mean and dividing by the standard deviation.
            After normalization, we randomly flip the energyscape along both the horizontal and vertical directions, before passing the (possibly flipped) energy scape to the CNN model as input.

    \subsection{Models}
    \label{subsec:methodology:models}
        \noindent Most of the models we use were taken from the Scikit-learn python library~\cite{Pedregosa2011Scikit}, with the exception of the gradient boosted trees, which were taken from the xgboost python library~\cite{Chen_2016_XGBoost}.
		The CNN was implemented using PyTorch~\cite{Paszke_2017_Automatic}.
    
        \subsubsection{CNN}
        \label{subsubsec:methodology:models:deep_learning}
            \noindent The CNN architecture is based on the overall skeleton laid out in NAS-Bench-101~\cite{Ying2019NASBench101}.
			Data is first fed into a convolutional stem, which increases the number of channels.
			Following this, the architecture consists of a sequence of residual blocks, with downsampling modules interspersed at regular intervals.
			Following this, global average pooling is used to reduce the feature maps to a single vector, which is then fed through a linear layer to compute the logits of our classifier network.
			All convolutions are executed using Convolution-BatchNorm-ReLU-Dropout order.
    
            \begin{figure}
                \centering
                \includegraphics[width=0.8\columnwidth]{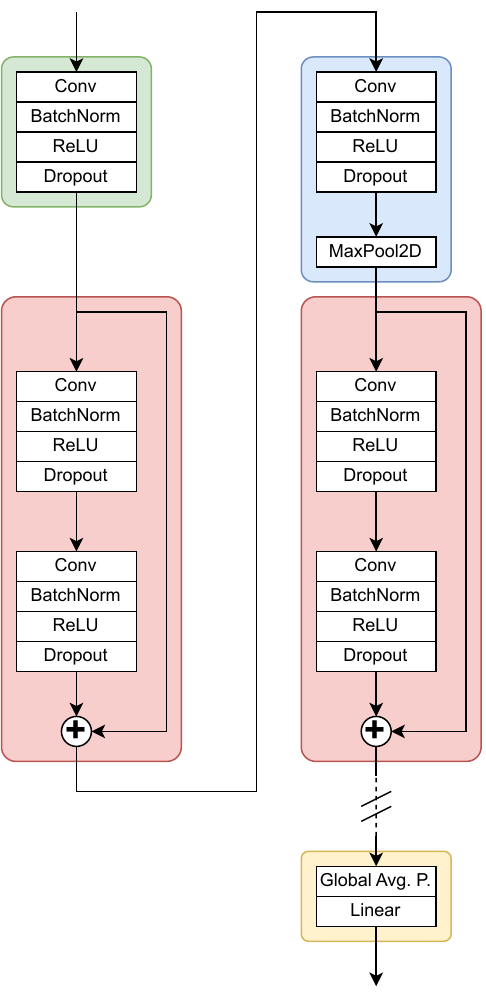}
                \caption{A schematic representation of the used neural network architecture. The stem convolution is marked in green in the top-left, the schematic shows two residual modules marked in red, and one downsampling module with a blue background in the top-right. Finally, the global average pooling and linear layers are included at the bottom-right.}
                \label{fig:cnn-architecture}
            \end{figure}
    
            Figure~\ref{fig:cnn-architecture} shows a schematic representation of the CNN architecture we used.
			The number of layers in the architecture is determined based on the uniform scaling principle introduced by EfficientNet~\cite{Mingxing2019Efficient}, with some slight modifications.
    
            Following~\cite{Mingxing2019Efficient}, the size of our architecture is determined by four hyperparameters: $\alpha, \beta, \gamma$ and $\phi$.
			From these, we compute three ratios: the depth ratio ($\alpha^{\phi}$), width ratio ($\beta^{\phi}$), and the resolution ratio ($\gamma^{\phi}$).
            We then take a set of baseline values, manually selected for the $\phi=0$ case, and multiply each of the ratios with these baseline values to obtain the number of layers (from the depth ratio, $d = d_{\phi=0} \cdot \alpha^{\phi}$), the initial number of channels (from the width ratio, $w = w_{\phi=0} \cdot \beta^{\phi}$) and the interval between downsampling layers (from the resolution ratio, $r = r_{\phi=0} \cdot \gamma^{\phi}$).
            The number of residual blocks is determined by the number of layers, minus one, to account for the stem convolution. After every ``downsampling interval'' residual blocks, we insert a downsampling module.
			A downsampling module consists of a Conv-Bn-ReLU-Dropout module followed by a pooling module.
			The convolution operation doubles the number of channels, while the pooling module halves the spatial resolution.
			Note that the downsampling interval affects both the number of channels and the spatial resolution of the featuremaps.
			The number of channels can be controlled individually using the width ratio. Thus, the downsampling interval primarily serves to control the spatial resolution of the feature maps.

\section{Experiments}
\label{sec:experiments}
	\noindent The hyperparameters we used were kept constant across all experiments.
	We use PyTorch Lightning {2.6.0}, PyTorch {2.10.0}, Numpy {2.4.1}, Scikit-learn {1.8.0} and XGBoost {3.1.3}.

    \subsection{Data collection parameters}
    \label{subsec:experiments:data_collection}
    	\noindent Some parameters of the dataset were set during data collection.
		Every SONAR sample in the dataset consists of a recording with 163840 samples sampled at a rate of 4.5MHz from all 32 microphones, encoded using PDM.
		The emitted signal by the sensor is a hyperbolic chirp with a frequency ranging from 20kHz to 50kHz. Thic chirp is generated with a sample rate of 450 kHz, and has a duration of 2.5ms.

	\subsection{Preprocessing parameters}
	\label{subsec:experiments:preprocessing}
		\noindent Since the preprocessing pipelines for the vector-based models differ from that of the CNN model, we discuss each pipeline separately here.
		The preprocessing of the SONAR data is done on the SONAR sensor using the on-board compute unit.
		This limits the possible computational complexity of the preprocessing pipeline, particularly affecting operations like beamforming etc., necessitating a trade-off between preprocessing complexity and sample rate.

		For both pipelines, beamforming is done in 91 directions, with both azimuth and elevation ranging from -90 to +90 degrees, essentially covering the full frontal hemisphere.

		\subsubsection{Vector preprocessing pipeline}
			\noindent Max-filtering the energyscapes to reduce their range-wise dimension is done with a kernel size of 5. The PCA analysis is configured to whiten the data first, before projecting down to 256 components.

		\subsubsection{Image preprocessing pipeline}
			\noindent For the CNN preprocessing pipeline, we time-shift the signals following PDM decoding and matched filtering by a random amount of samples, with a maximum of 45 samples.
			At this point in the pipeline, the sample rate of the signals is 450kHz, thus, a delay of 45 samples corresponds to a time delay of 100\textmu s.
			Following beamforming, envelope detection and clean-up, we flip the received energyscapes horizontally (along the azimuth/elevation dimension) and vertically (along the range dimension), with each flip occurring with a probability of 50\%.

    \subsection{Model Parameters}
    \label{subsec:experiments:models}
    	\noindent With the exception of the CNN model, all models were trained in a one-vs-rest setting to enable multi-label classification.
		Except when mentioned otherwise, we leave model parameters to their default values.

    	\subsubsection{Logistic Regression}
    	\label{subsubsec:experiments:models:logistic_regression}
	        \noindent As mentioned at the start of section~\ref{sec:methodology}, for each experiment, we make use of three different models: A logistic regression model, a support vector machine, a decision tree, a random forest, a multi-layer perceptron, a gradient-boosted tree and, for the damage detection problem, a CNN.
			For the logistic regression model, we set C (The inverse of the regularization strength) to 0.01, and used an L2 penalty for regularization.
			We fitted both slope and intercept. To mitigate issues caused by dataset imbalance, we utilized the ``class\_weight'' option in scikit-learn, by setting it to ``balance''.
			This has the effect of weighting the loss associated with each sample based on Equation~\ref{eq:sklearn_class_weight_balanced}, which weighs samples based on the number of occurrences that a sample belongs to.

	        \begin{equation}
	        	w_{i} = \frac{S}{C * \sum_{j = 0}^{S}{Y[j, i]}}
	        	\label{eq:sklearn_class_weight_balanced}
	        \end{equation}
	        In this equation, $w_{i}$ is the weight of the i-th class, the feature matrix, has a dimension of $S \times F$, where $S$ is the number of samples, and $F$ the number of features.
			$Y$ Is the label matrix, with a dimension of $S \times C$, where $C$ is the number of classes.

	        We performed each experiment twice, with the seed for random number generation set to 0 or 1, depending on the experiment to account for variations resulting from different initializations of the optimization algorithms.

        \subsubsection{Support Vector Machine}
        \label{subsubsec:experiments:models:svm}
	        \noindent For the Support Vector Machine (SVM)-based models, we set C (inversely related to the regularization strength) to 1, an L2 penalty is used for regularization, we use the same class weighting we used for the logistic regression, shown in equation~\ref{eq:sklearn_class_weight_balanced}.
			Our SVM uses a radial basis function (RBF) kernel.
			We set the $\gamma$ parameter in our RBF kernel to the value shown in equation~\ref{eq:sklearn_svm_gamma_scale}, this corresponds to the ``scale'' setting in Scikit-learn. 
	
			\begin{equation}
				\gamma = \frac{1}{F \cdot Var[X]}
				\label{eq:sklearn_svm_gamma_scale}
			\end{equation}
		
			The variables in Equation~\ref{eq:sklearn_svm_gamma_scale} correspond to those in Equation~\ref{eq:sklearn_class_weight_balanced}.
			Similar to the logistic regression setting, each experiment was performed twice. The SVMs were also trained in the same one-vs-rest setting as the logistic regression models.

		\subsubsection{Decision Tree}
			\noindent The parameters for the decision tree model were largely left to their default values in sci-kit learn, with the exception of the ``class\_weight'' parameter, which was set to ``balanced'', similar to the logistic regression model.
			Similar to the other models, decision trees were also trained in a one-vs-rest setting.

		\subsubsection{Random Forest}
			\noindent Similar to before, the random forest model's parameters were largely left to their default settings, with the exception of the ``class\_weight'' parameter, which was simiarly set to balanced.

		\subsubsection{Gradient-Boosted Trees}
			\noindent As we did with the random forest model, we left the parameters for the gradient-boosted tree models on their default settings.

		\subsubsection{Multi-Layer Perceptron}
			\noindent The multi-layer perceptron consists of an input layer, three hidden layers with 256 neurons each, and an output layer performing the classification.
			An L2 regularization was applied, using a weight of $1 \times 10^{-5}$, the network was trained in minibatches of 256 samples over 250 iterations with a fixed learning rate of 0.01 using the Adam optimizer~\cite{Kingma_2017_Adam}.

		\subsubsection{CNN}
		\label{subsubsec:experiments:models:deep_learning}
			\noindent Our CNN models were trained with a batch size set to 1024, for 1000 epochs, with a learning rate of 0.1.
			After 100 and 800 epochs of training, the learning rate was reduced to 0.01 and 0.001, respectively.
			Gradients were clipped to have a norm with a magnitude of 20, and accumulated over 4 batches before being applied.
			The neural network is regularized using an L2 weight decay multiplied by a factor of $1 \times 10^{-7}$.
			We used LeakyReLU activations everywhere instead of standard ReLU (Rectified Linear Unit) activations.
			Across the network, batchnorm was applied with a momentum of 0.1.
			The model size parameters $\alpha, \beta, \gamma$ and $\phi$ were set to $\alpha=1.2$, $\beta=1.1$, $\gamma=1.15$ and $\phi=2$.
			The model size parameters are set to $d_{\phi=0} = 3$, $w_{\phi=0} = 16$, and $r_{\phi=0} = 1$.
			The stem convolution had a kernel size of $7 \times 11$ and was applied with a stride of 3 by 5.
			Dropout was enabled in the stem convolution, with a probability of 0.1.
			Both convolution operations in a residual block were applied with a kernel size of $9 \times 5$.
			In the downsampling module, the convolution operation had a kernel size of $1 \times 1$, a stride of 1 in both dimensions, and uses no padding.
			Dropout was disabled in the downsampling module, while the max-pooling operation had a kernel size of $3 \times 3$, and was applied with a stride of 2, with zero padding in both dimensions.
	
			All experiments were carried out on a virtual machine with a 48 (virtual) CPU cores of an AMD EPYC 9654 CPU and an NVIDIA L40S GPU with 48GB of VRAM.

\section{Results}
\label{sec:results}
	\noindent In this section, we present the results obtained in the two different problem domains, by each of the models.
	We evaluated each of the models on 2 parameters: The obtained F1 score, and the obtained Cohen's Kappa coefficient, denoted as $\kappa$.
	The F1 and Cohen's Kappa scores were averaged across classes based on the number of true samples of each class.
	An in-depth discussion of these results can be found in Section~\ref{sec:discussion}.

    \subsection{Material Classification}
    	\noindent In Table~\ref{tab:material-classification-performance}, we showcase the performance of all different models on the material classification task.
		Each model was trained twice with a different random initialization, and was trained using 10-fold cross validation.
		Performance numbers were averaged across all folds and random initializations.
    
    	\begin{table*}
    		\begin{center}
				\caption{The results of the different models on the material classification problem, averaged across random initalizations and folds. All numbers are given as $\mu\pm\sigma$. The best result in each column is marked in bold.}
				\label{tab:material-classification-performance}
				\begin{tabular}{lllllll}
					\hline
					Model                  & Test $\kappa$               & Validation $\kappa$         & Training $\kappa$           & Test F1                     & Validation F1               & Training F1                 \\
					\hline
					Logistic Regression    & $40.34\%\pm1.22\%$          & $48.00\%\pm3.18\%$          & $45.18\%\pm0.68\%$          & $71.77\%\pm0.90\%$          & $75.73\%\pm1.73\%$          & $74.66\%\pm0.77\%$          \\
					Decision Tree          & $76.82\%\pm1.90\%$          & $92.03\%\pm1.23\%$          & $89.49\%\pm0.48\%$          & $89.03\%\pm0.63\%$          & $95.93\%\pm0.47\%$          & $94.88\%\pm0.16\%$          \\
					Random Forest          & $76.84\%\pm0.69\%$          & $93.25\%\pm1.46\%$          & $91.16\%\pm0.42\%$          & $89.44\%\pm0.26\%$          & $96.64\%\pm0.54\%$          & $95.82\%\pm0.19\%$          \\
					Support Vector Machine & $74.67\%\pm1.66\%$          & $89.97\%\pm2.59\%$          & $88.03\%\pm2.02\%$          & $88.18\%\pm1.71\%$          & $95.02\%\pm1.73\%$          & $94.28\%\pm1.74\%$          \\
					Multi-Layer Perceptron & $77.52\%\pm2.25\%$          & $91.95\%\pm2.21\%$          & $89.42\%\pm1.34\%$          & $89.12\%\pm1.18\%$          & $95.95\%\pm1.08\%$          & $95.03\%\pm0.76\%$          \\
					Gradient-Boosted Trees & $\mathbf{78.10\%\pm1.40\%}$ & $\mathbf{93.58\%\pm1.83\%}$ & $\mathbf{91.37\%\pm1.07\%}$ & $\mathbf{89.55\%\pm0.60\%}$ & $\mathbf{96.79\%\pm1.01\%}$ & $\mathbf{95.95\%\pm0.51\%}$ \\
					\hline
				\end{tabular}
			\end{center}
   		\end{table*}

    \subsection{Damage Detection}
    	\noindent Similar to the previous section~, we report the F1 and $\kappa$ score for all models, averaged across random intializations and dataset folds in Table~\ref{tab:damage-detection-performance}.
		As noted before, in the damage detection problem, we additionally considered a CNN (ResNet) model.

		\begin{table*}
			\begin{center}
				\caption{The results of the different models on the damage detection problem, averaged across random initalizations and folds. All numbers are given as $\mu\pm\sigma$. The best result in each column is marked in bold.}
				\label{tab:damage-detection-performance}
				\begin{tabular}{lllllll}
					\hline
					Model                  & Test $\kappa$               & Validation $\kappa$         & Training $\kappa$           & Test F1                     & Validation F1               & Training F1                 \\
					\hline
					Logistic Regression    & $25.88\%\pm1.03\%$          & $30.14\%\pm2.35\%$          & $27.88\%\pm1.25\%$          & $49.16\%\pm0.58\%$          & $52.69\%\pm1.17\%$          & $51.20\%\pm0.31\%$          \\
					Decision Tree          & $54.98\%\pm1.51\%$          & $65.23\%\pm2.36\%$          & $61.34\%\pm0.86\%$          & $68.42\%\pm0.78\%$          & $75.76\%\pm1.60\%$          & $73.18\%\pm0.31\%$          \\
					Random Forest          & $57.87\%\pm1.46\%$          & $65.96\%\pm2.03\%$          & $62.11\%\pm1.01\%$          & $70.69\%\pm0.69\%$          & $76.40\%\pm1.39\%$          & $73.85\%\pm0.49\%$          \\
					Support Vector Machine & $52.64\%\pm1.25\%$          & $59.90\%\pm1.63\%$          & $56.56\%\pm0.81\%$          & $68.33\%\pm1.01\%$          & $73.15\%\pm1.40\%$          & $70.94\%\pm0.91\%$          \\
					Multi-Layer Perceptron & $54.15\%\pm1.82\%$          & $63.54\%\pm2.15\%$          & $59.44\%\pm1.41\%$          & $69.25\%\pm0.95\%$          & $75.83\%\pm1.25\%$          & $73.22\%\pm0.62\%$          \\
					Gradient-Boosted Trees & $\mathbf{58.00\%\pm2.38\%}$ & $\mathbf{66.66\%\pm2.66\%}$ & $\mathbf{62.52\%\pm2.10\%}$ & $\mathbf{71.74\%\pm1.15\%}$ & $\mathbf{77.75\%\pm1.54\%}$ & $\mathbf{75.06\%\pm0.90\%}$ \\
					ResNet  			   & $2.57\%\pm1.67\%$			 & $5.50\%\pm2.18\%$		   & $19.70\%\pm4.16\%$			 & $28.68\%\pm8.03\%$		   & $31.59\%\pm8.35\%$			 & $50.30\%\pm2.08\%$ \\
					\hline
				\end{tabular}
			\end{center}
		\end{table*}

\section{Discussion}
\label{sec:discussion}
	\noindent We used the same set of machine learning models in both the surface material classification and damage detection settings.
	
	Comparing the results in Table~\ref{tab:material-classification-performance} and Table~\ref{tab:damage-detection-performance}, we can see some similar trends appearing.
	First, we notice that the logistic regression model does significantly worse than the other models on both problems.
	This is to be expected, given its linear nature, compared to the other non-linear models.
	
	Next, we also note that for most models, standard deviations in both $\kappa$ and F1 score tend to be quite low, ranging from 0.5\% up to 2\%, with the exception of the ResNet used for damage detection.
	The higher standard deviation observed with the ResNet results is likely a result of the overall worse performance, as well as the complicated training procedure.

	We also observe that while training and validation data tend to be quite close together, there is somewhat of a gap with the performance on the test set.
	The split between the test and training+validation set was made in a non-stratified fashion, using a uniform distribution across samples.
	This may indicate a bias in the sampling strategy, leading to a test and training+validation set with different label distributions, leading to a performance gap between both.
	An alternative explanation is that the overall machine learning pipeline, and the hyperparameters for the models were overfitted to the training+validation set.
	This seems unlikely for a number of reasons.
	First of all, the difference exists consistently across different models. Given the limited amount of hyperparameter tuning that was done for different models, it seems unlikely that all models would have overfit to the same degree.
	Additionally, the preprocessing pipeline used for the SONAR data also doesn't contain any steps that may benefit one type of damage over another.
	The difference is also present in both the CNN model, and the other models, which use different preprocessing pipelines, which makes a bias in the preprocessing pipeline a less likely cause.
	
	Across models and problems, we can see that the models tend to perform slightly better on the validation set than on the training set.
	This is to be expected, as the optimal classification threshold was determined using Youden's J-statistic on the validation set, rather than the training set, to encourage generalization.
	This classification threshold was then used to evalute on all datasets (training, validation and test).

	
	\subsection{Material Classification}
	\label{subsec:discussion:material-classification}
		\noindent In this section, we discuss the results on the material classification problem.
		
		When looking at the statistics in Table~\ref{tab:material-classification-performance}, we can see that most models obtained F1 scores on the test set approaching 90\%, with $\kappa$ scores approaching 80\%.
		This represents a near-perfect classification, and shows that determining the road surface type using ultrasound sensing is possible, in-line with existing literature, such as~\cite{Kim_2021_Road}.

		From the poor performance of the logistic regression model on this dataset, we can also conclude that a linear model lacks the necessary complexity to fully capture the non-linear relationship between PCA components derived from energyscapes, and the type of road surface present in a measurement.

		We note that there is a large gap between the $\kappa$ score obtained on the test set, and the $\kappa$ score obtained on the validation and training set.
		We surmise that this is likely the case because the optimal classification threshold for each model was determined based on the validation dataset, leading to a classifier that generalizes well from the training set to the validation set, but likely does not perform as well on the test set.

		We also note that the test set is notably smaller than the training set (513 samples versus 4386 samples).
		Since some classes are quite rare, it is likely they only occur once or twice in the test set.
		Thus, a single missed classification in one of these rarer classes is likely to heavily affect the $\kappa$ score for that class, and consequently for all classes on the test set, with this effect being much less pronounced on the training set, since a single sample represents a much smaller part of the training set.
		This further highlights the importance of gathering a sufficiently large dataset, both for training and evaluation to ensure the system can both be built and evaluated well.

\begin{figure*}
			\centering
			\includegraphics[width=0.8\textwidth]{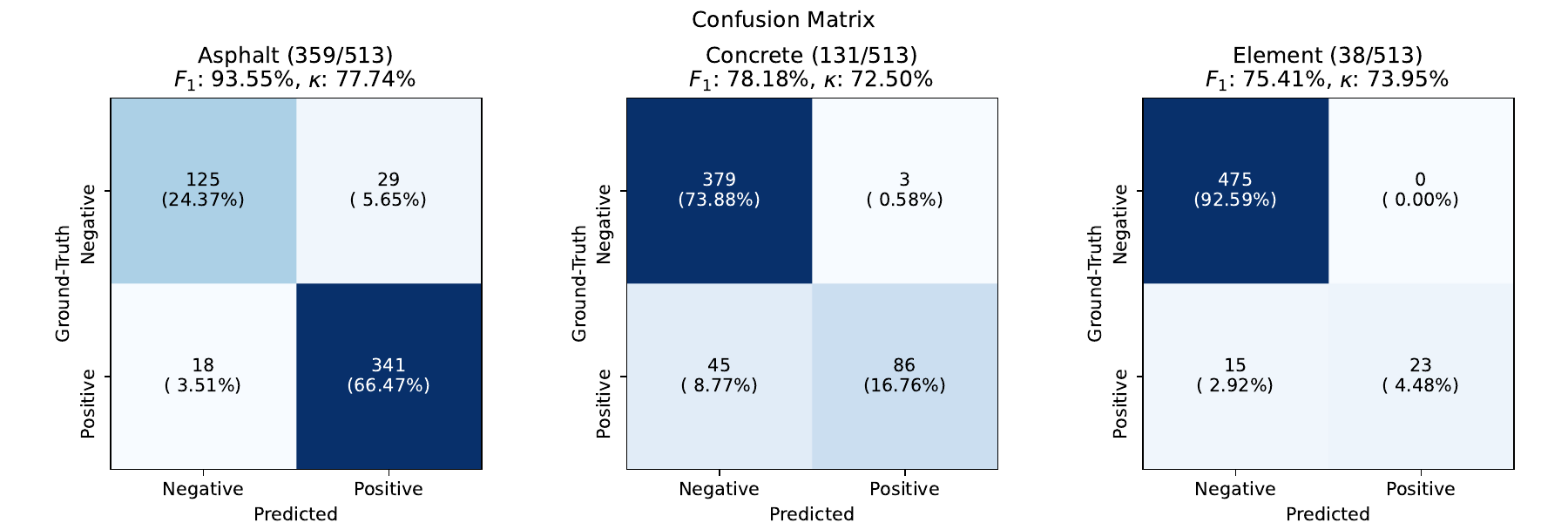}
			\caption{A confusion matrix of the test set for the gradient boosted trees classifier, trained on the first fold of the dataset for material classification, using a random seed of 0.}
			\label{fig:cm:xgb-materials-exp0-fold0}
		\end{figure*}

		Figure~\ref{fig:cm:xgb-materials-exp0-fold0} shows a confusion matrix for the gradient boosted trees model trained for material classification.

	\subsection{Damage Classification}
	\label{subsec:discussion:damage-classification}
		\noindent When looking at the damage classification results, we immediately see that the models performed significantly worse than on the material classification task.
		This is not unexpected, given the more subtle and complex nature of detecting and identifying road surface damages, compared to just identifying the surface type.
		Similar to before, we observe that the linear model does significantly worse than the others, clearly showing the need for a non-linear model.

		Counter to expectations, the CNN (ResNet) model did significantly worse than every other model, with the exception of the F1 score on the training set, where it barely managed to match a linear model in performance.
		We attribute this poor performance to the formatting of the image data fed to the CNN model. This is showcased in Figure 9 in~\cite{Steckel_2013_Broadband}.
		The figure shows interleaved areas of intensity for single reflections, due to the interleaved nature of the azimuth/elevation direction.
		Because azimuth and elevation are collapsed into a single matrix dimension, a discontinuity occurs whenever the minor index reaches its end value.
		This discontinuity violates an underlying assumption of the convolution operation: That adjacent samples are close in some sense of locality (temporal, spatial, etc.).
		It is likely this violated assumption that was responsible for the poor performance of the CNN model.

\begin{figure*}
			\centering
			\includegraphics[width=\textwidth]{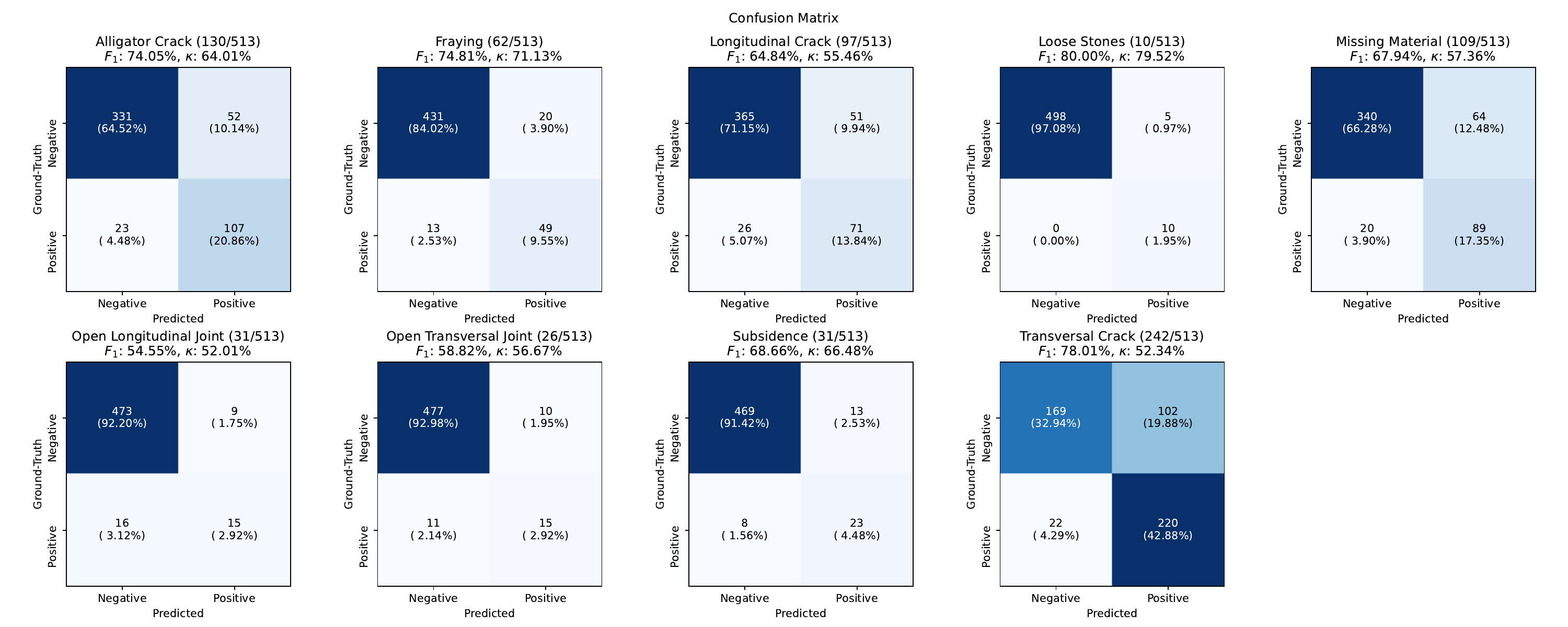}
			\caption{A confusion matrix of the test set for the gradient boosted trees classifier, trained on the first fold of the dataset for damage classification, using a random seed of 0.}
			\label{fig:cm:xgb-damages-exp0-fold0}
		\end{figure*}

		Figure~\ref{fig:cm:xgb-damages-exp0-fold0} shows a confusion matrix for the gradient boosted trees model trained for damage classification.

\section{Future Work}
\label{sec:future_work}
	\noindent When assembling the dataset used in this paper, location information was not used to inform the splitting of samples between the test, validation and training set.
	Given the limited size of the dataset, this would 	likely create further class imbalance, and make it even more difficult to obtain satisfactory classification performance.
	We should also note that, under imperfect data collection conditions, where different sensors can fail, the inclusion of a second (GPS) sensor can reduce the available number of samples, since these samples are the subset of all samples where both the SONAR and GPS sensors are operational.
	Despite this, the current method of splitting the dataset can result in data leakage, by samples taken close together in physical space, being split between the training, validation and test sets.
	A better approach would be to cluster samples based on the road they were taken from, and then splitting the roads between the different datasets, to ensure that all data from a single road ends up in the same set.
	This would improve the generalizibility of the models by ensuring that they also function appropriately on streets not seen before.
	
	
	In this work, we made use of a simple, linear delay-and-sum beamforming algorithm.
	While this works and has been used in the past succesfully in numerous applications~\cite{Steckel_2024_Tool,Jansen_2024_Semantic,Schenck_2025_Toward}, more advanced beamforming algorithms with better capabilities exist.
	Examples of this include the Minimum Variance Distortionless Response (MVDR) beamformer~\cite{Balasem_2012_Beamforming}, or the Delay-Multiply-and-Sum (DMAS)~\cite{Jansen_2025_Delay} beamforming algorithms.
	Such algorithms can results in cleaner data following the beamforming process, but also increase the computational load on the on-board compute units.
	
	To improve the perfomance of the convolutional neural networks, two possible solutions prevent themself.
	The first solution involves re-organizing the energyscapes into 3-dimensional tensors, and using 3-dimensional convolutions for the recognition of damages.
	While this would resolve the issues presented, it would also significantly increase the computational complexity of the neural network.
	Taking into account that the CNN is already significantly slower than the other models to train, this trade-off may not be worthwhile, with the CNN already taking several days to train, compared to several minutes the other models.
	Alternatively, the signal processing pipeline could be altered to support 2-dimensional beamforming along the elevation direction.
	While this would resolve the issue, it also has the disadvantage that some information is lost with regards to the size and location of the damages, which may be an informative feature for the detection system.

\section{Conclusion}
\label{sec:conclusion}
	\noindent In this paper, we showcased the viability of 3D in-air SONAR for the monitoring of road surfaces as part of a pavement management system.
	We show that 3D in-air SONAR sensing is a viable approach for detecting the material used for the road surface, with our results being in-line with the pre-existing results in the scientific literature.
	For damage detection, we show promising results, exceeding our own expectations, but falling short of what is required for an industrially viable solution.
	Based on these results, we argue that 3D in-air SONAR sensing is a worthwhile modality for inclusion in an opportunistic sensing system as part of a PMS, though further research is needed to improve damage detection performance.
	We highlight several avenues for further research, including the inclusion of additional sensor data, improved signal processing pipelines, and improved machine learning pipelines.




\section*{Funding}
    This work was realized in imec.ICON Hybrid AI for predictive road maintenance (HAIROAD) project, with the financial support of Flanders Innovation \& Entrepreneurship (VLAIO, project no. HBC.2023.0170).


\printbibliography














\end{document}